\documentclass[letterpaper, 10 pt, conference]{ieeeconf}  

\IEEEoverridecommandlockouts                              

\title{\LARGE \bf
Learning to See While Learning to Act: Diffusion Models for Active Perception in Robot Imitation}

\author{Kuancheng Wang$^{1}$\quad
        Vaibhav Saxena$^{1}$\quad
        Shuo Cheng$^{1}$\quad
        Yotto Koga$^{2}$\quad
        Danfei Xu$^{1,3}$\\[0.5em]
        {\normalsize $^{1}$Georgia Institute of Technology\quad
        $^{2}$Autodesk Research\quad
        $^{3}$NVIDIA}\\[0.5em]
        {\normalsize \faGlobe\ \url{https://see2act.github.io/}\qquad
        \faGithub\ \url{https://github.com/KuanchengWang/See2Act}}%
}

\usepackage[utf8]{inputenc}
\usepackage[T1]{fontenc}
\usepackage[english]{babel}
\usepackage{url}
\usepackage{amsfonts}
\usepackage{nicefrac}
\usepackage{microtype}
\usepackage{amsmath}
\usepackage{amssymb}
\usepackage{mathtools}
\usepackage{subcaption}
\usepackage{booktabs}
\usepackage{ragged2e}
\usepackage[table,dvipsnames]{xcolor}
\usepackage{etoolbox}
\usepackage{xspace}
\usepackage{xpatch}
\usepackage{enumerate}
\usepackage{xstring}
\usepackage{setspace}
\usepackage{tabularx}
\usepackage{graphicx}
\usepackage{stackengine}
\usepackage{pgffor}
\usepackage{hyperref}
\usepackage[capitalise,noabbrev,nameinlink]{cleveref}
\usepackage[useregional=numeric]{datetime2}
\usepackage{wrapfig}
\usepackage[export]{adjustbox}
\usepackage{ifthen}
\usepackage{pifont}
\usepackage{titlesec}
\usepackage[normalem]{ulem}

\usepackage{algorithm}
\usepackage{algpseudocode}
\usepackage{caption}
\usepackage{cuted}  
\usepackage{wrapfig}
\usepackage{xcolor}
\usepackage{xspace}
\usepackage{cuted}
\usepackage{caption}

\newcommand{\methodname}{See2Act\xspace}

\usepackage{algorithm}
\usepackage{algpseudocode}
\usepackage{float}
\let\labelindent\relax
\usepackage{enumitem}
\usepackage{fontawesome5}

\definecolor{mylightpurple}{RGB}{218, 161, 217}
\newcolumntype{Y}[1]{>{\centering\arraybackslash}p{#1}}
\definecolor{mypurple}{RGB}{191, 64, 191}
\begin{document}
\bstctlcite{IEEEexample:BSTcontrol}

\maketitle
\thispagestyle{empty}
\pagestyle{empty}


\begin{strip}
    \centering
    \includegraphics[width=\textwidth]{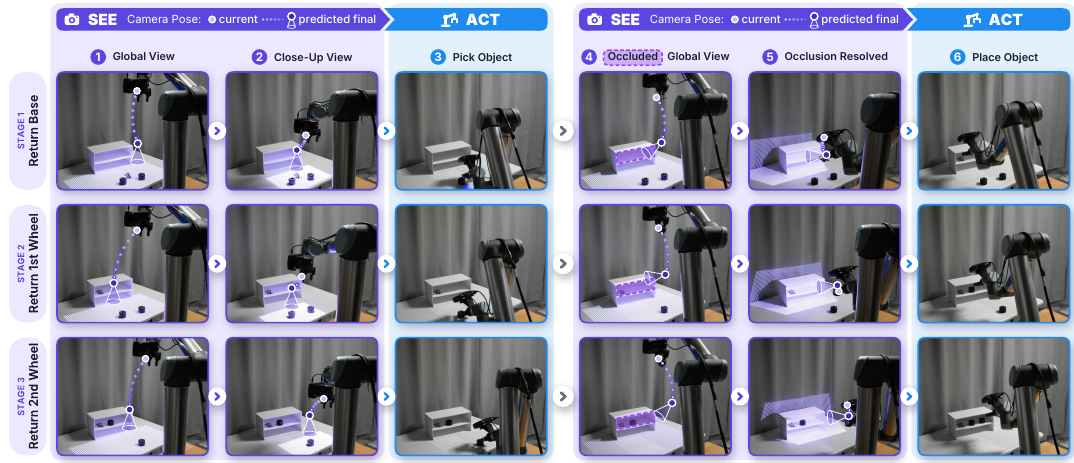}
    \captionof{figure}{\textbf{\methodname{} couples \emph{seeing} and \emph{acting} in a single denoising loop.} At each denoising step, the camera view and the predicted action are refined together: the action sets the next viewpoint (transition shown in white dotted trajectory), and the new viewpoint denoises the next action. This results in a policy that repositions the camera to reveal placement positions initially hidden from overhead views, achieving 95\% \emph{zero-shot} sim-to-real transfer on tasks with occlusion.}
    \label{fig:teaser}
    \vspace{-3mm}
\end{strip}
\begin{abstract}
Most imitation learning methods assume full observability in table-top settings. In practice, objects are often occluded, requiring robots to both \emph{search} and \emph{act}, and learning this coupled behavior from limited demonstrations remains challenging. We propose \methodname, an imitation learning approach that conditions action prediction on a sequence of actively-inferred viewpoints at test time, by coupling action denoising with viewpoint refinement. The policy is trained using camera poses anchored to keyframe actions from offline demonstrations, enabling implicit learning of where to \textit{see}, while learning how to \textit{act}. We empirically demonstrate that in Ravens the policy recovers informative viewpoints under severe occlusions, and on RLBench tasks it improves performance by up to 33\% points over prior methods. In the real world, we collect 50 demonstrations in a digital twin and achieve zero-shot sim-to-real transfer on pick-and-place tasks using depth observations. The policy handles significant occlusions, showing that learned viewpoint reasoning enables robust manipulation under partial observability.
\end{abstract}

\section{Introduction}
\label{sec:intro}
Imitation learning (IL) for visuomotor control~\cite{robomimic2021,saxena2025mimiclabs} typically assumes fixed viewpoints, conflating the abilities to \emph{see} and \emph{act}. In realistic settings, however, a robot must decide \emph{where to look}, especially under occlusion when task-relevant objects are not initially visible. This introduces a fundamental challenge: demonstrations entangle two distinct behaviors -- viewpoint selection and manipulation -- which require different forms of supervision. While manipulation can often be learned from aligned state-action pairs, selecting informative viewpoints requires spatial reasoning over unseen regions of the scene. Simply augmenting demonstrations with additional viewpoints increases data requirements combinatorially, without teaching the policy which views are useful. Consequently, policies trained with limited data often fail in occluded environments, revealing a key limitation of current IL approaches.

Recent methods have significantly improved data-efficient visuomotor control. Diffusion policies \cite{chi2024diffusionpolicyvisuomotorpolicy} model actions through conditional denoising, with further gains from 3D representations \cite{ze20243d}, symmetry-aware architectures \cite{wang2024equivariant,ryu2024diffusion}, and camera-conditioned refinement \cite{saxena2024c3dm}. Keyframe-based approaches \cite{pmlr-v205-shridhar23a, goyal2023rvt, goyal2024rvt} improve efficiency by supervising only task-critical waypoints. However, these methods remain passive in perception, assuming that the required visual information is already visible at inference time. As a result, policies with fixed or pre-defined viewpoints often fail under occlusion, where informative observations must be actively acquired. The missing capability is therefore not only data-efficient imitation learning, but policies that jointly learn \emph{where to see} and \emph{how to act}.

The central insight of this work is that manipulation should be treated as a coupled process of action refinement and visual evidence acquisition, rather than action prediction from a fixed viewpoint. We introduce \textbf{\methodname}, a diffusion-based imitation learning framework that jointly refines robot actions and camera viewpoints. At test time, a wrist-mounted camera iteratively updates its pose alongside the denoised 6-DoF action, actively acquiring observations that reduce task ambiguity. This behavior emerges from training the diffusion model on target actions paired with viewpoint trajectories anchored to the same demonstration keyframe, enabling the policy to learn which viewpoints are most informative for precise manipulation under occlusion. Training is performed entirely in simulation using scalable viewpoint-agnostic data generation, and the resulting policy transfers zero-shot to real robots using depth observations to reduce the sim-to-real gap. In summary, our key contributions are:
\begin{enumerate}
    \item We present \methodname, a viewpoint-agnostic imitation learning policy that learns viewpoint-action coupling by leveraging digital twins and a novel training strategy for keyframe-action prediction from offline manipulation datasets.

    \item \methodname enables a test-time mechanism \textit{Active Viewpoint Inference}, that iteratively refines the robot's camera pose and predicted action to autonomously resolve occlusions.

    \item We create three new tasks based on the Ravens Benchmark \cite{zeng2020transporter}, on which \methodname achieves 91\% average success rates while addressing the search-vs-manipulation learning disparity. We further evaluate on nine RLBench tasks \cite{james2020rlbench}, achieving an average success rate of 81.1\%.

    \item Finally, we demonstrate zero-shot sim-to-real transfer on two real-world tasks with explicit occlusions, achieving average success rate of 95\% without any real-world fine-tuning.
\end{enumerate}

\section{Related Work}
\label{sec:related}

\noindent\textbf{Diffusion models for visuomotor policies.} Diffusion Policy~\cite{chi2024diffusionpolicyvisuomotorpolicy} formulates visuomotor control as conditional denoising over action trajectories, enabling multimodal action generation. Subsequent works improve spatial generalization using point clouds~\cite{ze20243d}, SO(2) symmetry~\cite{wang2024equivariant}, or SE(3) bi-equivariance~\cite{ryu2024diffusion}, often relying on privileged geometric information. C3DM~\cite{saxena2024c3dm} additionally couples denoising with fixation-based zooming into task-relevant regions, but limits camera motion to translational zoom along a fixed orientation. \methodname{} extends this to full 6-DoF camera trajectories, allowing active reorientation and repositioning to uncover occluded regions.

\noindent\textbf{Keyframe actions in robotics.} Keyframe-based action representations decompose trajectories into critical waypoints, reducing supervision while focusing on decisive task moments~\cite{zawalski2024robotic}. Prior methods include voxelized point-cloud approaches such as PerAct~\cite{pmlr-v205-shridhar23a}, view-based methods RVT/RVT-2~\cite{goyal2023rvt,goyal2024rvt}, 3D feature-field prediction in Act3D~\cite{gervet2023act3d}, and enhanced spatial-temporal encoding in PolarNet and HiveFormer~\cite{pmlr-v229-chen23b,guhur2022instructiondrivenhistoryawarepoliciesrobotic}. However, all rely on fixed camera configurations and cannot actively acquire new views when task-relevant regions are occluded.

\textbf{Active perception.} Choosing where to sense is a classical problem: next-best-view methods pick the sensor pose that most reduces scene uncertainty~\cite{Connolly1985TheDO}, and interactive perception uses the robot's own actions to disambiguate what it sees~\cite{Bohg_2017}, with perception and manipulation kept as separate stages. Recent learned methods control an external camera from data, imitating human active perception with a robotic neck~\cite{xiong2025visionactionlearningactive}, learning gaze by reinforcement learning~\cite{kerr2025eyerobotlearninglook}, or moving the camera to keep the task-relevant region observable while a separate policy acts~\cite{wang2025observeactasynchronousactive}. In all of these the camera is an independently actuated sensor and viewpoint selection is a module of its own, trained with a separate objective, reward, or planner. Multi-view picking~\cite{morrison2019multiviewpickingnextbestviewreaching} moves a wrist camera along the reach, but selects views with a hand-crafted grasp-entropy criterion. \methodname{} differs in both respects: the camera is the wrist camera of the manipulating arm, so views are acquired along the approach to the action, and each viewpoint is a deterministic function of the current action estimate, so the view sequence is supervised by the action labels alone and emerges from denoising without viewpoint annotations, rewards, or a planner.

\section{Method}
\label{sec:method_v2}
\begin{figure*}[tbp]
    \centering
    \resizebox{1.0\textwidth}{!}{\includegraphics{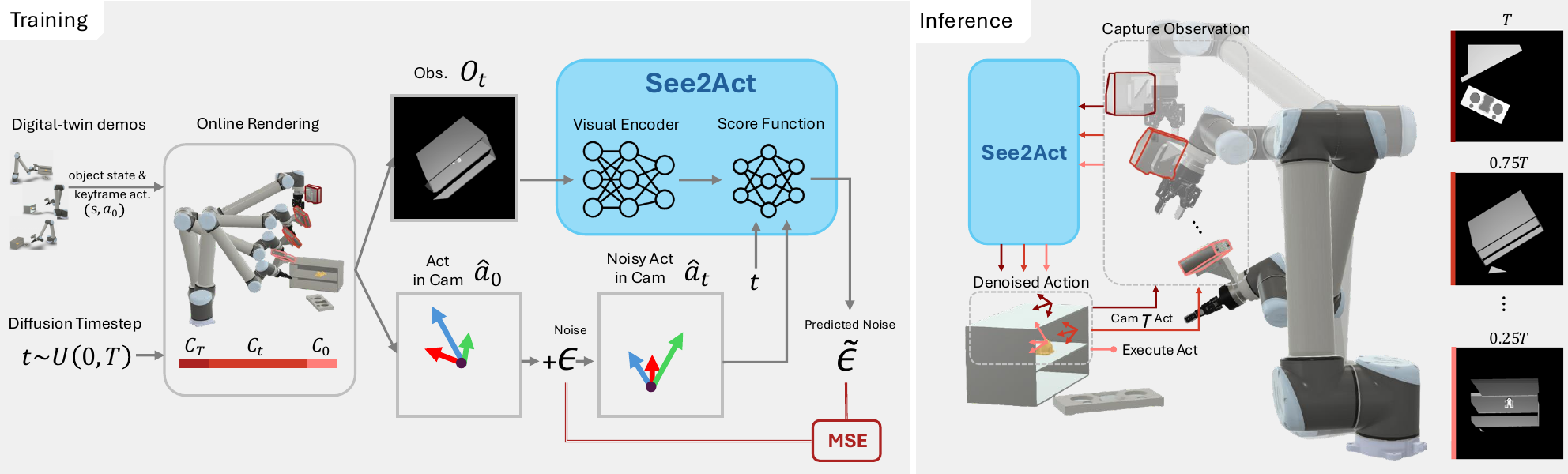}

    }
    \caption{\textbf{Training and Inference in \methodname.} 
    \textit{(Left: Training)} Demos are collected in a digital twin, which provides the object states $s$ and keyframe actions $a_0$. We sample diffusion timesteps $t$, compute camera poses $C_t$, render observations $O_t$ in the twin, and generate noisy actions $\hat a_t$ by transforming $a_0$ into each camera frame and adding noise. The visual encoder and noise predictor are trained jointly with an MSE loss on the predicted noise $\tilde\epsilon$. \textit{(Right: Inference)} Starting from an overview ($t=T$), \methodname{} iteratively refines both the action and camera pose. At each denoising step, it captures an observation, predicts noise, updates the action estimate, and computes the next camera pose. The final action is executed by the robot.
    }
    \label{fig:method_fig}
    \vspace{-3mm}
\end{figure*}
Visuomotor policies with fixed viewpoints are limited by what a single observation reveals. We
address this by embedding active perception directly into the diffusion process. The key insight
behind \methodname{} is that the diffusion trajectory is also a trajectory over viewpoints. During
training, each noisy action is paired with a camera pose and observation anchored to the same
target keyframe: early diffusion steps see broad contextual views and later steps see progressively
target-centric ones. At inference, denoising the action updates the camera pose, so the policy
acquires observations increasingly informative for the action it is predicting, turning passive
denoising into active perception without a separate viewpoint planner or reward. Figure~\ref{fig:method_fig} summarizes both processes.

\subsection{Extracting and Noising Keyframe Actions}
We build on DDPMs \cite{ho2020denoising} to model the conditional distribution over target actions given image observations. We represent actions in the camera frame rather than a fixed world frame: the camera pose defines the coordinate frame in which the action is expressed, so denoising the action and moving the camera become two views of one trajectory. 
This coupling enables the policy to jointly reason about \textit{where to see} and \textit{how to act} throughout the denoising process. 

We assume each demonstration is segmented into keyframes, as defined in~\cite{pmlr-v205-shridhar23a}. The keyframe action $a_0$ is the target end-effector pose at the upcoming keyframe, $a_0 \triangleq (p_{a,0}, \theta_{a,0}) \in \mathbb{R}^6$, with position $p_{a,0} \in \mathbb{R}^3$ and Euler-angle orientation $\theta_{a,0} \in \mathbb{R}^3$.

To realize the forward diffusion process, we first express the target action in the frame of the camera pose $C_t = (p_{c,t}, \theta_{c,t})$ of timestep $t$ (Section~\ref{sec:method_b}),
\begin{equation}
\hat a_0 = \mathcal{T}_{C_t}(a_0) \triangleq \big( R(\theta_{c,t})^{\top}(p_{a,0} - p_{c,t}),\; \mathcal{E}(R(\theta_{c,t})^{\top} R(\theta_{a,0})) \big),
\label{eq:to_cam}
\end{equation}
where $R(\theta) \in SO(3)$ is the rotation matrix of Euler angles $\theta$ and $\mathcal{E}(\cdot)$ returns the Euler angles of a rotation. Although $a_0$ is fixed in the world frame, $\hat a_0$ is viewpoint-dependent. We then noise $\hat a_0$ with a predefined schedule $\bar\alpha(t)$,
\begin{equation}
\hat a_t = \hat a_0 + \sqrt{1 - \bar\alpha(t)}\,\epsilon, \qquad \epsilon \sim \mathcal{N}(0, I),\; t \in [0, T].
\label{eq:forward}
\end{equation}
Note that we use the noising process {without drift} as proposed in \cite{saxena2024c3dm} due to its better performance empirically for denoising diffusion models with changing contexts.
\subsection{Generating Camera Poses for Policy Conditioning}
\label{sec:method_b}
\begin{wrapfigure}{r}{0.23\textwidth}
    \vspace{-0.2cm}
    \centering
    \includegraphics[width=0.23\textwidth]{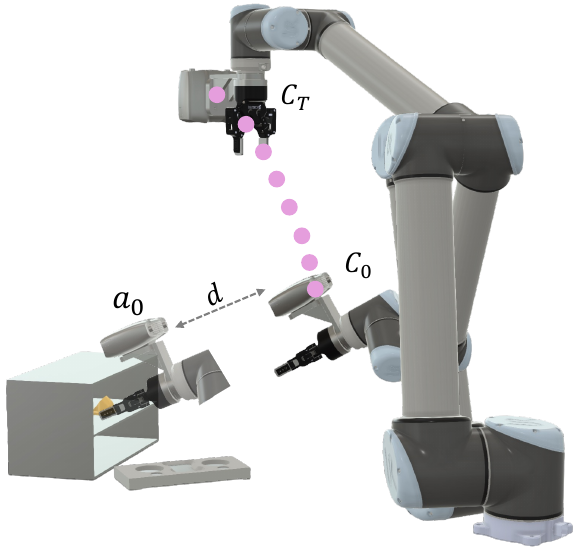}
    \caption{\textbf{Camera Pose Generation}, where the \textcolor{mylightpurple}{purple} dots show the interpolation for $C_t$.}
    \label{fig:cam_traj_generation}
    \vspace{-0.3cm}
\end{wrapfigure}
We associate each diffusion timestep with a camera pose: the denoising trajectory hence also creates a trajectory of observations, running from a broad global view at timestep $T$ to a target-centric view at timestep 0, which we obtain in simulation. The trajectory is anchored at two poses: $C_T$ and $C_0$ at diffusion timesteps $T$ and $0$ respectively. $C_T\triangleq(p_{c,T},\theta_{c,T})$, is a fixed pose of our choice that gives a broad view of the scene for context, while $C_0=(p_{c,0},\theta_{c,0})$ is derived from the target keyframe action $a_0$ and gives a close, target-centric view (Figure~\ref{fig:cam_traj_generation}).

We construct $C_0$ by placing the camera at a fixed offset from the target action and pointing it back along the action's approach axis, so the final view is centered on the action frame. Specifically, we have:
\begin{equation}
p_{c,0} = p_{a,0} + d\,R(\theta_{a,0})\,\mathbf{z}, \qquad R(\theta_{c,0}) = R(\theta_{a,0})\,R_x(\pi),
\label{cam_target_position_orientation}
\end{equation}
where $\mathbf{z} \triangleq [0,0,1]^{\top}$ is the approach axis, $d$ the standoff distance, and $R_x(\pi)$ a half-turn about the camera's $x$ axis, which points the camera (imaging along its $+z$ axis) back at the target. Since $C_0$ is a function of the keyframe action alone, the camera trajectory is supervised by the action labels, with no viewpoint annotations. Intermediate poses interpolate the anchors with $\tau \triangleq t/T \in [0,1]$, positions linearly,
\begin{equation}
p_{c,t} = (1 - \tau)\,p_{c,0} + \tau\,p_{c,T},
\label{intermidate_cam_position}
\end{equation}
and orientations along the shortest arc between the anchor quaternions $q_{c,0} = q(\theta_{c,0})$ and $q_{c,T} = q(\theta_{c,T})$, with signs chosen so that $q_{c,0} \cdot q_{c,T} \ge 0$,
\begin{equation}
\theta_{c,t} = \mathcal{E}\Big( \tfrac{\sin((1-\tau)\Omega)}{\sin\Omega}\, q_{c,0} + \tfrac{\sin(\tau\Omega)}{\sin\Omega}\, q_{c,T} \Big)
\label{intermidate_cam_orientation}
\end{equation}
where $\Omega = \arccos(q_{c,0} \cdot q_{c,T})$. Interpolating smoothly in both position and orientation matters for the policy, not just for visual smoothness: it gives a continuous sequence of viewpoints, each paired with a progressively less noisy action estimate, so neighboring denoising steps see observations that change gradually rather than drastically jumping between unrelated views.
Since both position and orientation vary smoothly, the resulting camera trajectory is continuous in $SE(3)$ and can be followed on a physical arm. Along it, the camera moves from a broad observation toward a target-centric one anchored by the action, which brings occluded or fine-grained task regions into view as denoising proceeds.
\subsection{Learning Image-Conditioned Action Denoising}
Because our method requires observations from multiple viewpoints during training, we assume access to a dataset collected in a digital twin $\mathcal{D} = \{ (s^{(i)}, {a}^{(i)}_0) \}_{i=1}^{N}$, where $s^{(i)}$ denotes the initial object state for demonstration $i$, and $a^{(i)}_0$ denotes the corresponding target action. We additionally assume an online renderer that generates an observation $O_t$ by initializing the scene to state $s$ and rendering an image from camera pose $C_t$. 
The renderer needs the geometry of the task objects and their poses $s$, not photorealistic appearance: on the real robot the policy consumes depth images (Section~\ref{sec:real_robot_exp}), so the twin has to match shape and layout only, and any view synthesizer with the interface $(s, C_t) \mapsto O_t$, including a reconstruction from a real capture, can take its place.
Our model consists of a shared visual encoder $e_\phi$ and a noise prediction network $\epsilon_\psi$ associated with each keyframe action. Given a dataset of size $N$ and $K$ latent action samples generated per demonstration, we optimize the standard mean-squared error objective:
\begin{equation}
\mathcal{L} \triangleq \frac{1}{NK} \sum_{i=1}^{N} \sum_{k=1}^{K} \big\| \epsilon_\psi\big(e_\phi(O^{(i,k)}_{t}), \hat a^{(i,k)}_{t}, t^{(i,k)}\big) - \epsilon^{(i,k)} \big\|^{2},
\end{equation}
with $t^{(i,k)} \sim \mathcal{U}(0,T)$ and $\epsilon^{(i,k)} \sim \mathcal{N}(0,I)$ drawn per sample, $O^{(i,k)}_t$ rendered from $s^{(i)}$ at $C_t$, and $\hat a^{(i,k)}_t$ from Eqs.~\eqref{eq:to_cam}--\eqref{eq:forward}.
\subsection{Active Viewpoint Inference}
\label{sec:active_viewpoint_inference}
At inference time, we jointly execute the reverse diffusion process and the camera schedule, leveraging the coupling learned during training to continuously reposition the camera as the action is denoised. We refer to this procedure as \emph{active viewpoint inference}, and use $\tilde{\cdot}$ throughout this subsection to mark inferred quantities, distinguishing them from the ground-truth $C_t$, $a_t$ used during training. The world-frame estimate is initialized as $\tilde a_T \sim \mathcal{N}(0, I)$. At each step $t = T, \dots, 1$: (i)~the camera moves to $\tilde C_t$, obtained from $\tilde a_t$ by Eqs.~\eqref{cam_target_position_orientation}--\eqref{intermidate_cam_orientation} ($C_T$ at $\tau = 1$), and captures $\tilde O_t$; (ii)~the estimate is expressed in the camera frame and, for $t < T$, re-noised to level $t$, $\hat{\tilde a}_t = \mathcal{T}_{\tilde C_t}(\tilde a_t) + \sqrt{1-\bar\alpha(t)}\,\epsilon$; (iii)~from $\hat\epsilon_t = \epsilon_\psi(e_\phi(\tilde O_t), \hat{\tilde a}_t, t)$, inverting Eq.~\eqref{eq:forward} gives $\hat{\tilde a}_0 = \hat{\tilde a}_t - \sqrt{1-\bar\alpha(t)}\,\hat\epsilon_t$; (iv)~$\tilde a_{t-1} = \mathcal{T}^{-1}_{\tilde C_t}(\hat{\tilde a}_0)$ returns to the world frame and sets $\tilde C_{t-1}$. Since $\tilde C_{t-1}$ is conditioned on the current action estimate, improved action predictions lead to more target-centric viewpoints, which in turn produce observations better suited to refining the action estimate at subsequent steps. When several refinement runs are affordable, we run inference $n_{\text{rollout}}$ times, record the variance of the inferred camera-pose changes of each run, and keep the $\tilde a_0$ of the run with the smallest variance, i.e.\ the most stable inferred camera trajectory ($n_{\text{rollout}} = 1$ unless stated otherwise).

Iterating this procedure produces a coupled sequence of observation-action pairs, $(O_T, \tilde a_T) \rightarrow (\tilde O_{T-1}, \tilde a_{T-1}) \rightarrow \dots \rightarrow (\tilde O_1, \tilde a_1) \rightarrow \tilde a_0$, where the camera trajectory emerges from denoising, and $\tilde a_0$ is the final keyframe action. Active perception is thereby built into the diffusion process, allowing the policy to recover from an initially occluded view by steering the camera toward task-relevant regions without explicit viewpoint planning or reward.

\section{Experiments}
\label{sec:experiments_v2}
We evaluate \methodname{} across simulated and real-world manipulation tasks, organizing our experiments around a central claim: coupling viewpoint refinement to action denoising resolves partial observability in a way that fixed views, passive or planned view selection, restricted or closed-loop camera motion, and keyframe iteration without diffusion cannot. Hence, we ask:
\begin{enumerate}
\item Can active viewpoint refinement in \methodname{} resolve occlusions that fixed-view, passive-selection, zoom-only, closed-loop, and view-planning baselines cannot?
\item Which ingredients matter: does performance depend on the camera-frame coupling, on the diffusion head, and on the number of refinement steps?
\item Does \methodname{} remain competitive against strong multi-view and 3D manipulation policies on standard benchmarks?
\item Does the learned viewpoint behavior produce systematic search, and how sensitive is it to the initial camera pose $C_T$?
\item Does \methodname{} transfer zero-shot from digital-twin training to real-robot manipulation?
\end{enumerate}

\subsection{Simulation Experiments}

\subsubsection{Tasks and Baselines}

\begin{figure}[t]
\centering
\includegraphics[width=\columnwidth]{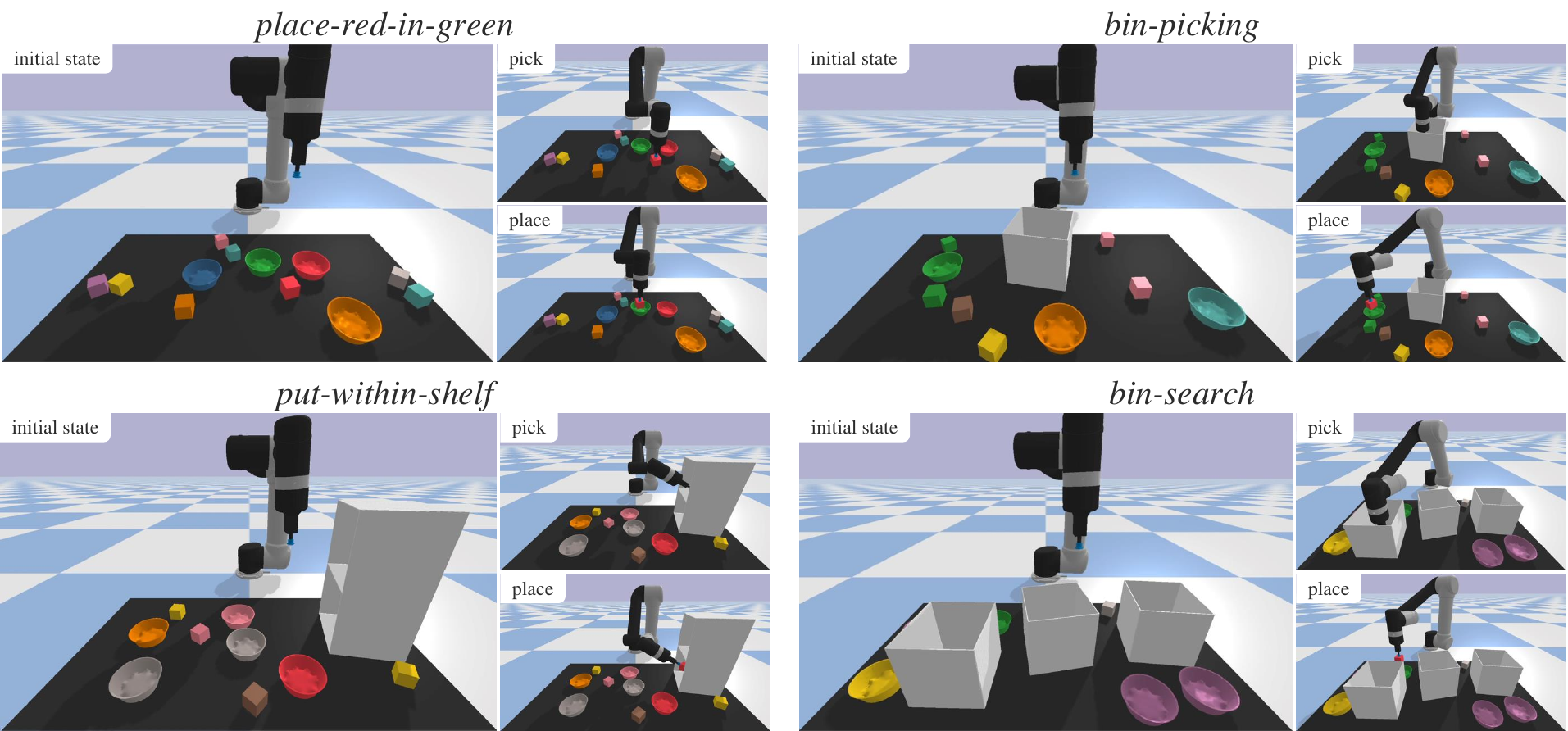}
\caption{\textbf{Ravens}~\cite{zeng2020transporter} tasks with occlusion; showing initial, pick, and place states.}
\label{fig:ravens_task}
\vspace{-6mm}
\end{figure}

\textbf{Ravens.} We use Ravens~\cite{zeng2020transporter} as a diagnostic benchmark for occlusion and search, with four tasks (Figure~\ref{fig:ravens_task}). The existing \textit{place-red-in-green} requires picking a red block and placing it in a green bowl among distractor blocks and bowls. Three new variants add \textit{front-view occlusions} of increasing difficulty: in \textit{bin-picking} the red block starts inside an open-topped bin and is invisible from the front view; in \textit{put-within-shelf} the block and the bowl are hidden on the top and bottom levels of a shelf; in \textit{bin-search} the block is placed at random inside one of three open-topped bins, while the bowl remains visible on the table.

We compare against nine baselines that share \methodname{}'s visual backbone and differ only in how views are obtained and used. (1)~\textbf{Conv-MLP} is behavioral cloning from a single \emph{fixed view}. (2)~\textbf{Conv-MLP-MV} and (3)~\textbf{Diff-MV} use \emph{fixed multi-view} observations with a regression and a diffusion action head, respectively (Diff-MV conditions on four fixed views). (4)~\textbf{Diff-MV-Entropy} adds \emph{passive view selection}: trained on randomly sampled views, it picks at test time, among ten preset views, the action with the lowest denoising entropy. (5)~\textbf{C3DM}~\cite{saxena2024c3dm} allows \emph{restricted camera motion}, zooming at a fixed orientation. (6)~\textbf{DP-Wrist} is a closed-loop Diffusion Policy~\cite{chi2024diffusionpolicyvisuomotorpolicy} on the wrist camera, trained on the same demonstrations; it re-observes at every control step but never reasons about where to look. (7)~\textbf{Diff-NBV} and (8)~\textbf{Diff-IG} implement \emph{classical active perception}: \methodname{}'s diffusion head, trained to denoise the action from a single random view given the camera pose, is driven by a separate planner that moves the camera step by step under the same budget of 20 views per keyframe. At each step the planner scores candidate views around the current action estimate; Diff-NBV moves to the view where the denoised actions agree most (uncertainty-driven next-best-view), and Diff-IG, reading the denoised actions as hypotheses of the target location, to the view that would observe the most of them (information gain). 
(9)~\textbf{Keyframe-Iter} keeps \methodname{}'s viewpoint loop but replaces camera-frame diffusion with a regression head that predicts the keyframe action directly in the world frame: predict, move the camera toward the prediction (Eqs.~\eqref{cam_target_position_orientation}--\eqref{intermidate_cam_orientation}), and re-predict, for 20 iterations.
Together, the baselines test whether occlusion can be resolved by fixed views, by passive or classical view selection, by restricted or closed-loop camera motion, or by keyframe iteration without diffusion, isolating active viewpoint refinement as the source of \methodname{}'s gains.

\textbf{RLBench.} We evaluate on nine RLBench~\cite{james2020rlbench} tasks spanning categorical reasoning (\textit{meat-off-grill}, \textit{slide-block}, \textit{sort-shape}), placement into occluded containers (\textit{put-in-cupboard}, \textit{put-in-safe}), high-precision insertion and placement (\textit{insert-peg}, \textit{place-wine}), and articulated-object manipulation (\textit{open-drawer}, \textit{turn-tap}).
We compare against Image-BC (CNN and ViT)~\cite{jang2022bc,pmlr-v205-shridhar23a}, C2F-ARM-BC~\cite{james2020rlbench}, HiveFormer~\cite{guhur2022instructiondrivenhistoryawarepoliciesrobotic}, PolarNet~\cite{pmlr-v229-chen23b}, PerAct~\cite{pmlr-v205-shridhar23a}, Act3D~\cite{gervet2023act3d}, RVT~\cite{goyal2023rvt}, and RVT-2~\cite{goyal2024rvt}, including several multi-view 3D policies with full point-cloud or multi-camera input. This allows us to test whether \methodname{}, using only a single 2D view per step, can match or outperform methods with substantially richer spatial information.

\textbf{Implementation and Training Details.} For each task in simulation, we collect 100 demonstrations from an oracle policy. We parameterize the visual encoder $e_\phi$ with a ResNet-18 backbone~\cite{he2016deep} using FiLM
conditioning~\cite{perez2018film} to incorporate language instructions, and implement the noise predictor $\epsilon_\psi$ as a fully-connected MLP with ReLU activations~\cite{agarap2018deep}. All baselines listed in Ravens share this backbone to ensure a fair comparison. We train with
Adam~\cite{kingma2014adam} at a learning rate of $10^{-4}$ and batch size 50 on a single NVIDIA RTX 3090 GPU. For RLBench~\cite{james2020rlbench}, we train task-specific policies rather than a single multi-task policy.

\subsubsection{Main results}
\label{ravens_results}
\begin{figure}[t]
\centering
\includegraphics[width=\columnwidth]{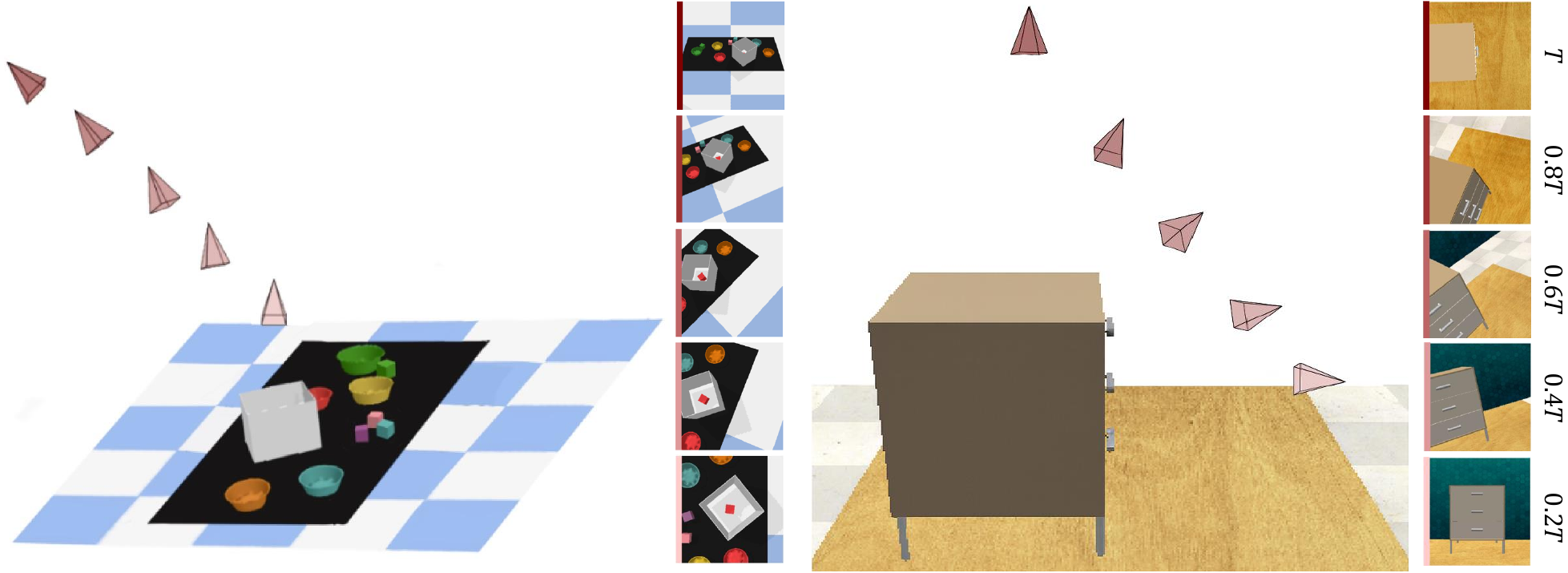}
\caption{\textbf{Active Viewpoint Inference.} Showing camera poses (pyramids) and observations. \methodname iteratively repositions the camera to reveal the occluded red block to pick for \textit{bin-picking} (left) and the drawer handle for \textit{open-drawer} (right).}
\label{fig:bin_picking_cam_traj}
\vspace{-3mm}
\end{figure}
\textbf{Active perception lets the policy look around occlusions.} 
In \textit{bin-picking} (Figure~\ref{fig:bin_picking_cam_traj}, left) active viewpoint inference generates camera poses starting from the distant global view ($t=T$),
where the red block is fully occluded, through intermediate views that circumvent occlusion, to a close-up at $t=0.2T$ that supplies fine-grained visual detail needed for the pick action. 

\textbf{\methodname{} beats methods with fixed-views (single or multiple) in the presence of occlusion.} 
Table~\ref{tab:ravens_table} shows that Conv-MLP, which tests if fixed-view BC suffices, fails across all occluded tasks. Conv-MLP-MV and Diff-MV, which test if multiple \emph{fixed} viewpoints are sufficient without view selection, also fail. Performance gains by \methodname{} indicate that simply supplying more pre-configured views does not resolve occlusion in challenging scenarios.


\begin{table}[t]
\centering
\small
\setlength{\tabcolsep}{3pt}
\begin{tabularx}{\columnwidth}{l
  >{\centering\arraybackslash}X
  >{\centering\arraybackslash}X
  >{\centering\arraybackslash}X
  >{\centering\arraybackslash}X}
\toprule
\textbf{Method}
& \textbf{\shortstack{\textit{place-red-}\\\textit{in-green}}}
& \textbf{\shortstack{\textit{bin-}\\\textit{picking}}}
& \textbf{\shortstack{\textit{put-within-}\\\textit{shelf}}}
& \textbf{\shortstack{\textit{bin-}\\\textit{search}}} \\
\midrule
Conv-MLP                     & 0  & 0  & 0  & 0 \\
Conv-MLP-MV                  & 0  & 2  & 0  & 0 \\
Diff-MV                      & 0  & 0  & 0  & 0 \\
Diff-MV-Entropy              & 54 & 22 & 2  & 64 \\
C3DM~\cite{saxena2024c3dm}   & 88 & 8  & 0  & 0 \\
DP-Wrist                     & 14 & 18 & 64 & 8 \\
Diff-NBV                     & 14 & 22 & 30 & 24 \\
Diff-IG                      & 12 & 18 & 30 & 26 \\
Keyframe-Iter                 & 18 & 22 & 46 & 12 \\
\textbf{\methodname\ (ours)} & \textbf{96} & \textbf{96} & \textbf{72} & \textbf{100} \\
\bottomrule
\end{tabularx}
\caption{Success rate (\%) on the Ravens~\cite{zeng2020transporter} tasks with occlusion, averaged over 50 trials per task.}
\label{tab:ravens_table}
\vspace{-4mm}
\end{table}
\textbf{\textit{Active} viewpoint refinement is necessary for manipulation under occlusion.} C3DM~\cite{saxena2024c3dm}, which allows only zoom-based camera motion, achieves 8\% on \textit{bin-picking} and fails on \textit{put-within-shelf} and \textit{bin-search}, indicating that translational repositioning alone cannot resolve occlusions requiring a change in viewing direction. Passive viewpoint selection with Diff-MV-Entropy reaches 22\% on \textit{bin-picking} but only 2\% on \textit{put-within-shelf}, likely because sampled views may not reveal the occluded region and cannot be refined during denoising. 
\begin{table*}[!t]
\centering
\small
\resizebox{1.0\textwidth}{!}{%
\setlength{\tabcolsep}{6pt}
\begin{tabular*}{\textwidth}{@{\extracolsep{\fill}}@{}lccccc@{}}
\toprule
\textbf{Models} & \textbf{Avg. SR $\uparrow$}  & \textbf{Insert-Peg} & \textbf{Meat-off-Grill} & \textbf{Open-Drawer} & \textbf{Place-Wine} \\
\midrule
Image-BC (CNN)~\cite{jang2022bc}  & 1.8 & 0 & 0 & 4 & 0 \\
Image-BC (ViT)~\cite{pmlr-v205-shridhar23a} & 1.8 & 0 & 0 & 0 & 0 \\
C2F-ARM-BC~\cite{james2020rlbench} & 17.3 & 4 & 20 & 20 & 8 \\
HiveFormer~\cite{guhur2022instructiondrivenhistoryawarepoliciesrobotic} & 54.7 & 0.0 & \textbf{100.0} & 52.0 & 80 \\
PolarNet~\cite{pmlr-v229-chen23b} & 52.4 & 4.0 & \textbf{100.0} & 84.0 &  40 \\
PerAct~\cite{pmlr-v205-shridhar23a} & 55.5 & 5.6 $\pm$ 4.1 & 70.4 $\pm$ 2.0 & 88.0 $\pm$ 5.7 & 44.8 $\pm$ 7.8 \\
Act3D~\cite{gervet2023act3d} & 70.6 & 27.0 & 94.0 & \textbf{93.0} & 80 \\
RVT ~\cite{goyal2023rvt}& 68.2 & 11.2 $\pm$ 3.0 & 88.0 $\pm$ 2.5 & 71.2 $\pm$ 6.9 &  91.0 $\pm$ 5.2 \\
RVT-2 ~\cite{goyal2024rvt}& 77.3 & \textbf{40.0} $\pm$ 0.0 & 99.0 $\pm$ 1.7 & 74.0 $\pm$ 11.8 & 95.0 $\pm$ 3.3 \\
\textbf{\methodname (ours)} & \textbf{81.1} & 24.0 $\pm$ 0.0& 98.0 $\pm$ 2.3 & 92.0 $\pm$ 0.0 & \textbf{99.0} $\pm$ 2.0\\
\midrule
\textbf{Models}  & \textbf{Put-in-Cupboard} & \textbf{Put-in-Safe} & \textbf{Slide-Block} & \textbf{Sort-Shape} & \textbf{Turn-Tap} \\
\midrule
Image-BC (CNN)~\cite{jang2022bc}  & 0 & 4 & 0 & 0 & 8 \\
Image-BC (ViT)~\cite{pmlr-v205-shridhar23a} & 0 & 0 & 0 & 0 & 16 \\
C2F-ARM-BC~\cite{james2020rlbench} & 0 & 12 &  16 & 8 & 68 \\
HiveFormer~\cite{guhur2022instructiondrivenhistoryawarepoliciesrobotic} & 32.0 & 76.0 & 64.0 & 8.0 & 80 \\
PolarNet~\cite{pmlr-v229-chen23b} & 12.0 & 84.0 & 56.0 & 12.0 & 80 \\
PerAct~\cite{pmlr-v205-shridhar23a} & 28.0 $\pm$ 4.4 & 84.0 $\pm$ 3.6 & 74.0 $\pm$ 13.0 & 16.8 $\pm$ 4.7 &  88.0 $\pm$ 4.4 \\
Act3D~\cite{gervet2023act3d}& 51.0 & 95.0 & 93.0 & 8.0 & 94 \\
RVT~\cite{goyal2023rvt}& 49.6 $\pm$ 3.2 & 91.2 $\pm$ 3.0 & 81.6 $\pm$ 5.4 & 36.0 $\pm$ 2.5 & 93.6 $\pm$ 4.1 \\
RVT-2~\cite{goyal2024rvt}& \textbf{66.0} $\pm$ 4.5 & \textbf{96.0} $\pm$ 2.8 & 92.0 $\pm$ 2.8 & 35.0 $\pm$ 7.1  & \textbf{99.0} $\pm$ 1.7 \\
\textbf{\methodname (ours)} & 60.0 $\pm$ 5.7 & \textbf{96.0} $\pm$ {0.0} & \textbf{100.0} $\pm$ 0.0 & \textbf{69.0} $\pm$ 2.0 & 92.0 $\pm$ 0.0 \\
\bottomrule
\end{tabular*}
} 
\caption{Success rate (\%) on RLBench~\cite{james2020rlbench}. We evaluate \methodname four times on each task with RLBench convention.}
\label{tab:rvt-results}
\vspace{-4mm}
\end{table*}
\textbf{Re-observing, planning views, or iterating keyframe predictions does not substitute for coupling.}
DP-Wrist re-observes at every control step yet reaches only 18\% and 8\% on
\textit{bin-picking} and \textit{bin-search}: the block stays hidden from the wrist camera as the arm approaches. It is competitive only on \textit{put-within-shelf} (64\% vs.\ 72\%),
whose fixed approach geometry suits a receding-horizon policy.
Classical view planning does not close the gap either: with the same budget of 20 views per keyframe,
Diff-NBV and Diff-IG reach at most 14\%, 22\%, 30\%, and 26\% on the four tasks. The bottleneck is the
decoupled policy, not the planner: even from the ground-truth terminal view $C_0$ it succeeds in only
8--24\% of trials, because a policy that predicts world-frame actions cannot exploit a close-up.
Keyframe-Iter, which keeps the viewpoint loop but regresses world-frame keyframe actions without diffusion, reaches
18\%, 22\%, 46\%, and 12\%. The converse ablation agrees: replacing the interpolation of
Eqs.~\eqref{intermidate_cam_position}--\eqref{intermidate_cam_orientation} with these classical rules inside
\methodname{}'s own denoising loop lowers success by up to 84\% points.
Overall, \methodname{} outperforms the Conv-MLP variants by up to 100\% points, Diff-MV-Entropy by up to 74\% points, C3DM by up to 100\% points , and the strongest new baseline on each task by 8--78\% points. What resolves occlusion is therefore
not more views, closer views, or planned views, but views coupled, in the camera frame, to the action being
denoised.

\begin{wrapfigure}{r}{0.25\textwidth}
\vspace{-1em}
    \centering
    \includegraphics[width=0.25\textwidth]{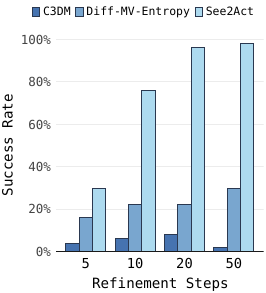}
    \caption{Success rates vs. refinement steps.}
    \label{fig:refinement_step}
\vspace{-2mm}
\end{wrapfigure}
\textbf{\methodname performance scales with finer camera refinements.} 
Figure~\ref{fig:refinement_step} examines how success rate scales with number of denoising steps on \textit{bin-picking}. \methodname improves drastically from 5 to 20 steps, while Diff-MV-Entropy and C3DM show marginal gains. This clearly shows that each viewpoint refinement adds significant value to action prediction, contrasting minimal gains with iterative model inference with action refinement only.  

\textbf{Both camera-frame coupling and diffusion are necessary.}
We remove each ingredient while keeping the same views, demonstrations and camera
schedule. Expressing the action in the world frame instead of the camera frame drops success to 26\%, 46\%, 52\%, 22\%; replacing the diffusion head by a regression head drops it to 74\%, 24\%, 4\%, 40\% on \textit{place-red-in-green}, \textit{bin-picking}, \textit{put-within-shelf}, \textit{bin-search} tasks respectively with the largest losses where the target is hidden from the initial view.

\textbf{\methodname{} remains competitive on diverse manipulation tasks.}\label{rlbench-results}
Beyond the occluded Ravens tasks, we evaluate active viewpoint refinement on standard RLBench tasks against methods with richer spatial input. RVT~\cite{goyal2023rvt} and RVT-2~\cite{goyal2024rvt} fuse observations from multiple pre-defined viewpoints, whereas \methodname{} uses a single actively repositioned view at each denoising step (Figure~\ref{fig:bin_picking_cam_traj}, right). Table~\ref{tab:rvt-results} shows that \methodname{} achieves the highest average success rate (81.1\%) and the best results on \textit{place-wine}, \textit{put-in-safe}, 
\begin{figure}[t]
    \centering
    \includegraphics[width=\columnwidth]{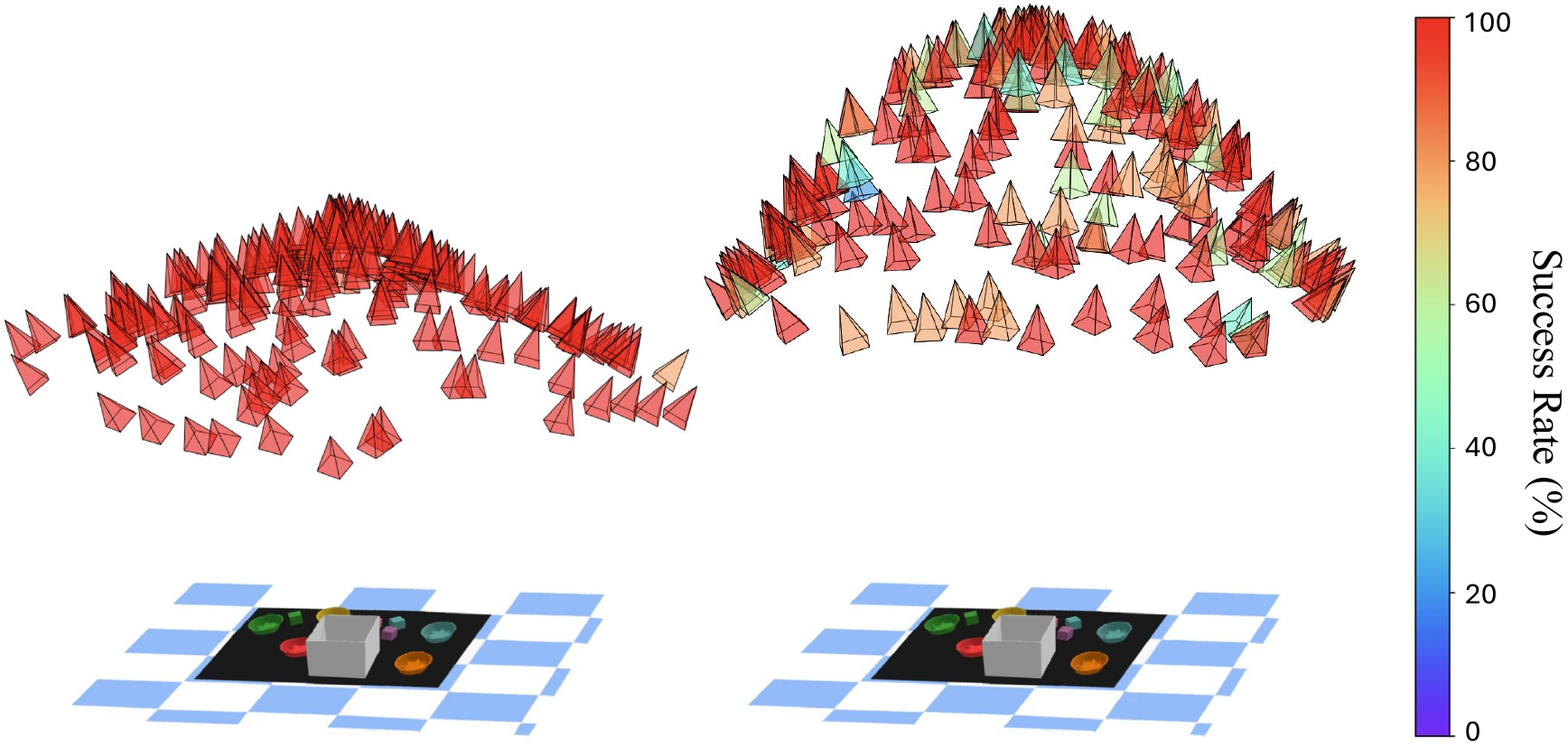}
    \caption{\textbf{Robustness to initial camera viewpoints.} Success rate associated with viewpoints sampled from in-distribution (left) and out-of-distribution (right).}
    \label{fig:cam_pyramid}
    \vspace{-6mm}
\end{figure}
\textit{slide-block}, and \textit{sort-shape}. The largest gain is on \textit{sort-shape}, suggesting that close-up views acquired through repositioning provide geometric detail unavailable from fixed fused views. These results indicate that active viewpoint refinement remains beneficial beyond occlusion-specific tasks and that a single actively-controlled 2D view can match or outperform point-cloud and multi-camera policies.

\textbf{\methodname{} demonstrates robustness to out-of-distribution initial camera viewpoints.}
We isolate how much \methodname{}'s performance depends on the initial camera pose $C_T$. We train the model
with $C_T$ sampled from a pyramidal region around the workspace
(Figure~\ref{fig:cam_pyramid}, left), exposing the policy to diverse starting perspectives.
Evaluated on in-distribution poses across five object configurations, \methodname{} reaches 97.5\%
on \textit{place-red-in-green} and 95.6\% on \textit{bin-picking}. On 200 out-of-distribution poses sampled from a larger dome-shaped region (Figure~\ref{fig:cam_pyramid}, right) that contains and extends beyond the training pyramid, performance drops  to 86.7\% and 88.2\%. This is consistent with the interpretation that active viewpoint refinement during
denoising partially compensates for a suboptimal starting view by steering the camera toward more
informative perspectives.

\textbf{Active perception shows emergent search behavior.} The \textit{bin-search} task tests whether search emerges from active viewpoint inference despite no explicit search supervision. With $n_{\text{rollout}}=10$, we observe abrupt camera reorientations when observations contradict the current action estimate (e.g., a selected bin is found empty), causing the next denoising step to inspect alternative candidates. Selecting the rollout with the smallest pose-change variance, \methodname{} achieves 100\% success on \textit{bin-search}, while 
fixed-view and zoom-only baselines score 0\% and the best view-planning baseline 26\%.
This suggests that iterative active refinement naturally induces search-like exploration that fixed-view and passive-selection methods cannot perform.

\subsection{Real-robot Experiments}
\label{sec:real_robot_exp}
We test whether: (1) a policy trained on 50 digital-twin demonstrations transfers zero-shot to a physical robot under occlusion, and (2) the active-perception mechanism in \methodname{} is deployable on a real robot. 

\textbf{Setup.} We use a UR10 arm, with a Robotiq gripper, and a wrist-mounted Zivid M60 depth camera. A digital twin replicates the hardware and real-world objects. 
Object models come from CAD models. All 50 demonstrations, their object poses $s$ and all training views are generated in the twin; at test time the policy uses only the wrist depth camera, with no real-world data, fine-tuning, or object-state estimation.
Camera calibration between simulation and reality is performed once before evaluation and kept fixed across all trials. We assume an off-the-shelf  motion planner and report only trials in which the planned trajectory is successful.

\begin{figure}[t]
    \centering
    \includegraphics[width=\columnwidth]{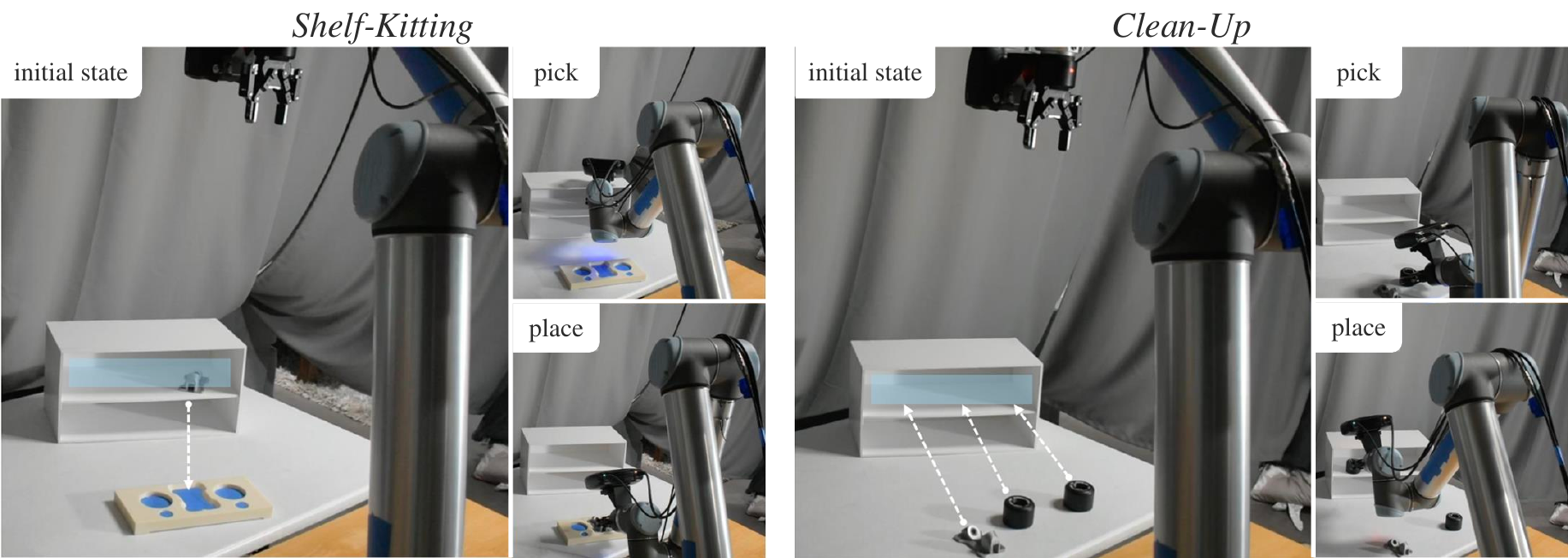}
    \caption{\textbf{Real-world tasks} showing initial, pick, place states. Blue region shows the area occluded from the camera at its initial pose. Arrows indicate transition from initial to goal.}
    \label{fig:real_task}
    \vspace{-2mm}
\end{figure}

\textbf{Tasks and Baselines.} We evaluate on 2 real-world tasks (Figure~\ref{fig:real_task}), involving occlusions from the initial overhead view. \textit{Shelf-kitting} is a high-precision insertion task in which a \textit{base} object, initially hidden inside a shelf, must be retrieved and inserted into a matching kit receptacle with 1--2 mm positional and 1--2$^\circ$ rotational tolerance; success is defined by releasing the base within this tolerance. The \textit{clean-up} task requires the robot to return two \textit{wheels} and one \textit{base} from randomized table locations to an occluded shelf, requiring camera repositioning to identify valid placement regions. To remove pick-order ambiguity, training labels correspond to the object closest to the camera-view center at each step.
We compare against Diff-MV and Diff-MV-Entropy, the fixed-view and passive-selection counterparts of \methodname{} sharing its backbone, twin data and depth pipeline; closed-loop and view-planning baselines are compared in simulation (Table~\ref{tab:ravens_table}).

\textbf{Training and Implementation Details.} \methodname{} is trained on 50 demonstrations per task, collected from 50 object layouts in the twin. To avoid online rendering during training, we pre-render 100 camera poses per demonstration along the trajectory prescribed by Section~\ref{sec:method_b} and apply image augmentation (cropping, rotation) during training. 
Trained on the same 50 demonstrations, the baselines (Diff-MV, Diff-MV-Entropy) fail on both tasks (Table~\ref{tab:pyatk_table}): a fixed-view policy maps global views of each layout to world-frame actions and must see many layouts before it generalizes, whereas \methodname{}'s target-centric views and camera-frame actions make different layouts look alike, so 50 suffice. We therefore also train the baselines on $3{\times}10^5$ layouts rendered in the twin, the regime in which they generalize. Since all training data are synthetic, the extra data cost only rendering time, and the asymmetry favors the baselines. \methodname{} needs only 50 demonstrations because its camera trajectory is generated from the action at inference rather than memorized from training. Training uses the same architecture and optimization settings as in simulation, on a single NVIDIA RTX 4090. 

\subsubsection{Results}


\begin{table}[t]
\centering
\small
\setlength{\tabcolsep}{4pt}
\begin{tabular*}{\columnwidth}{@{\extracolsep{\fill}}lccc@{}}
\toprule
\textbf{Models} & $\mathbf{N}_\text{demos}$ & \textbf{\textit{shelf-kitting}} & \textbf{\textit{clean-up}} \\
\midrule
Diff-MV & 50 & 0 & 0 \\
Diff-MV-Entropy & 50 & 0 & 0 \\
Diff-MV                       & $3\times10^{5}$ & 25 & 50 \\
Diff-MV-Entropy               & $3\times10^{5}$ & 50 & 50 \\
\textbf{\methodname{} (ours)} & \textbf{50}     & \textbf{95} & \textbf{95} \\
\bottomrule
\end{tabular*}
\caption{Real-world success rate (\%), and number of demos used for training.}
\label{tab:pyatk_table}
\vspace{-6mm}
\end{table}
\textbf{Active view refinement is essential at the precision required for tight-tolerance
insertion.} On \textit{shelf-kitting}, the gap between \methodname{} (95\%) and Diff-MV
(25\%) is largest (see Table~\ref{tab:pyatk_table}). Diff-MV's fixed overhead views lack the resolution needed for high-precision, whereas \methodname{}'s camera moves toward the target as
the action denoises, so the final denoising steps operate on close-up observations of the
receptacle. The result is consistent with the interpretation that close-up observations
acquired via active refinement carry geometric detail that fixed distant views cannot.

\textbf{Passive view selection helps but does not close the gap.} Diff-MV-Entropy (50\% on
both tasks) outperforms Diff-MV on \textit{shelf-kitting}, consistent with the view that selecting the most
informative view from a candidate set is preferable to fusing potentially mismatched ones.
But it cannot iteratively reposition, so it is bounded by the best view available in its
candidate set, which on these tasks is rarely close enough for high-precision placement.

\textbf{Active viewpoint inference improves robustness in sim-to-real transfer.} Diff-MV is sensitive to camera-pose discrepancies between sim and real, as its fused features become misaligned when the physical camera deviates from the fixed training viewpoints. Diff-MV-Entropy improves robustness (50\% vs.\ 25\%) by selecting a single low-entropy view instead of fusing misaligned observations, but its inability to reposition still limits performance. In contrast, \methodname{} is less affected by calibration error because the camera moves close to the target during denoising, so small angular offsets produce only small positional errors. Fixed-view methods must observe from farther away to cover the workspace, amplifying the same calibration errors.
\section{Limitations and Conclusion}
\label{sec:conclusion}
We presented \methodname, a diffusion-based visuomotor imitation learning framework that jointly denoises actions and full 6-DoF camera trajectories, enabling active perception to resolve occlusions that fixed-view and zoom-only methods cannot handle. Our approach achieves strong performance on tasks requiring active search and precise placement, and transfers zero-shot from simulation to a real robot using depth observations, reaching 95\% success with only 50 demonstrations collected on the physical twin. \methodname{} acquires a new observation at every denoising step, so execution is bounded by camera latency and slower than fixed-view methods. Training requires a digital twin with geometric models of the task objects, and inference relies on a motion planner to realize each camera pose on the arm. Extending \methodname{} from keyframe actions to dense trajectory prediction, for contact-rich tasks, is promising future work.


\bibliographystyle{IEEEtran}
\bibliography{references}
\end{document}